\documentclass[letterpaper, 10 pt, conference]{ieeeconf}
\IEEEoverridecommandlockouts
\usepackage{times}
\usepackage{epsfig}
\usepackage{graphicx}
\usepackage{amsmath}
\usepackage{amssymb}
\usepackage{booktabs}
\usepackage{pifont}
\usepackage{blindtext}
\usepackage{multirow}
\usepackage[table,xcdraw,rgb]{xcolor}
\usepackage{tikz}

\usepackage{fancyhdr}

\usepackage[breaklinks=true,bookmarks=false]{hyperref}

\title{\LARGE \bf
A Little Bit Attention Is All You Need for Person Re-Identification%
}

\makeatletter
\newcommand*\titleheader[1]{\gdef\@titleheader{#1}}
\AtBeginDocument{%
  \let\st@red@title\@title
  \def\@title{%
    \bgroup\normalfont\large\centering\@titleheader\par\egroup
    \vskip1.5em\st@red@title}
}
\makeatother

\titleheader{\phantom{.}\\[-10mm]\textit{\footnotesize Published as a conference paper at IEEE International Conference on Robotics and Automation (ICRA) 2023}\\[5mm]}

\author{Markus Eisenbach, Jannik L\"ubberstedt, Dustin Aganian, and Horst-Michael Gross%
\thanks{This work has received funding from the Carl-Zeiss-Stiftung as part of the project engineering for smart manufacturing (E4SM)}
\thanks{All authors are with Neuroinformatics and Cognitive Robotics Lab, TU Ilmenau, 98693 Ilmenau, Germany~~
{\tt\scriptsize markus.eisenbach@tu-ilmenau.de}
}}

\begin{document}

\maketitle
\thispagestyle{empty}
\pagestyle{empty}

\renewcommand*{\figureautorefname}{Fig.}
\renewcommand*{\tableautorefname}{Tab.}
\renewcommand*{\equationautorefname}{Eq.}
\renewcommand*{\sectionautorefname}{Sec.}
\renewcommand*{\subsectionautorefname}{Sec.}

\begin{abstract}
Person re-identification plays a key role in applications where a mobile robot needs to track its users over a long period of time, even if they are partially unobserved for some time, in order to follow them or be available on demand.
In this context, deep-learning based real-time feature extraction on a mobile robot is often performed on special-purpose devices whose computational resources are shared for multiple tasks.
Therefore, the inference speed has to be taken into account.
In contrast, person re-identification is often improved by architectural changes that come at the cost of significantly slowing down inference.
Attention blocks are one such example.
We will show that some well-performing attention blocks used in the state of the art are subject to inference costs that are far too high to justify their use for mobile robotic applications.
As a consequence, we propose an attention block that only slightly affects the inference speed while keeping up with much deeper networks or more complex attention blocks in terms of re-identification accuracy.
We perform extensive neural architecture search to derive rules at which locations this attention block should be integrated into the architecture in order to achieve the best trade-off between speed and accuracy.
Finally, we confirm that the best performing configuration on a re-identification benchmark also performs well on an indoor robotic dataset.
\end{abstract}

\section{Introduction}
In recent years, mobile robots that follow a user have become increasingly important, especially in the field of clinical rehabilitation.
One example is a robot coach for walking and orientation training of stroke patients~\cite{gross2017roreas}, where the robot had to accompany the patients during their walking exercises that they had to perform on their own to improve their mobility as well as their orientation skills.
This way, the robot addressed the patients' insecurity and anxiety of not being able to perform the exercises or not being able to find the way back to their apartment, which are possible reasons for neglecting self-training.
Another example is a robotic assistant that coached patients during their walking exercises, which were taught to them by physiotherapists after hip endoprosthetics surgery~\cite{scheidig2021robot,vorndran2018always}.
The robot was intended as a kind of physiotherapist replacement to provide immediate feedback to the patient regarding any deviations from the expected physiological gait pattern.

\begin{figure}[t!]
    \centering
    \includegraphics[width=\linewidth]{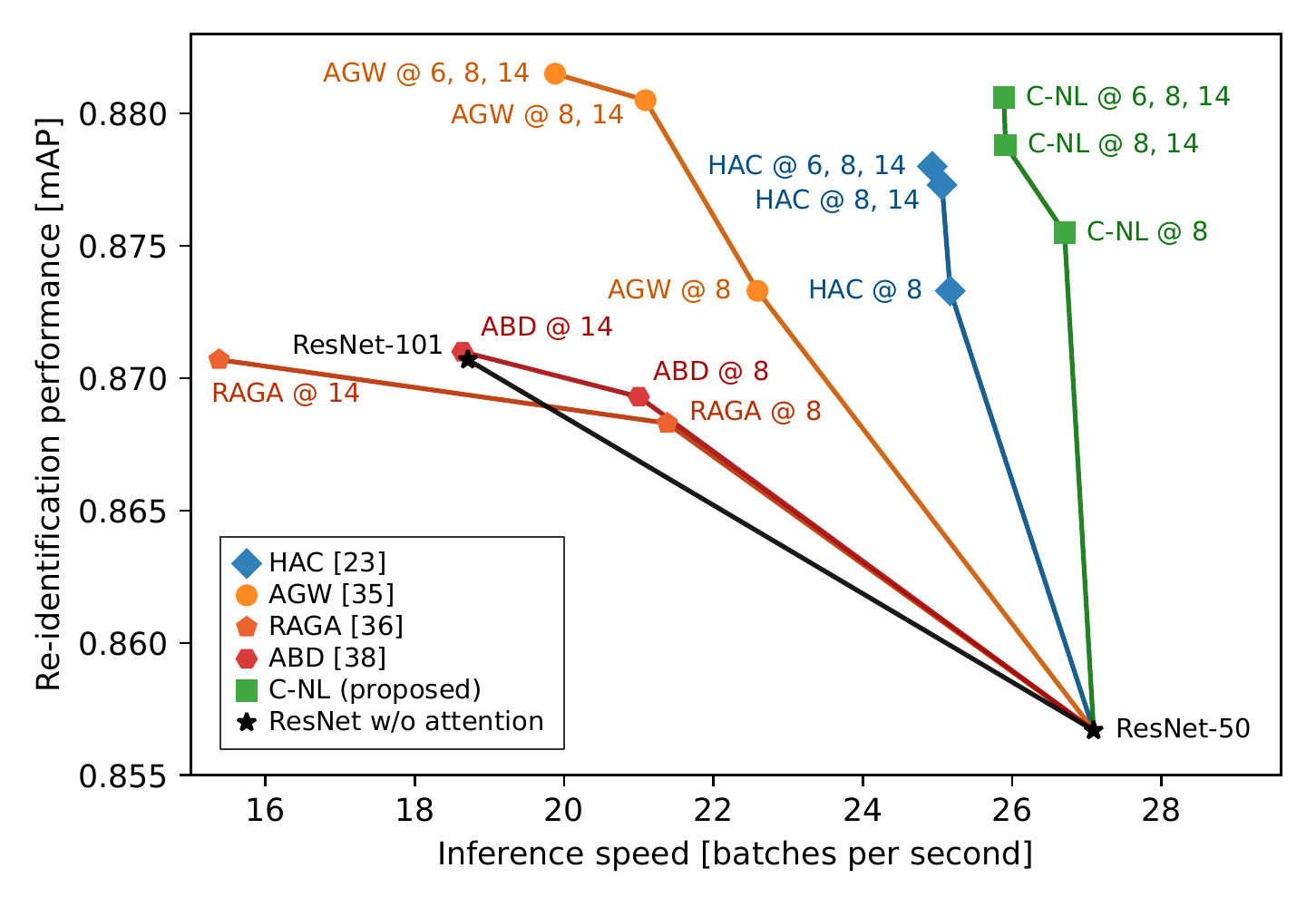}
    \caption{\label{fig:mapVsSpeed}Re-identification performance on the Market-1501 benchmark dataset \cite{zheng2015scalable} versus inference speed for different attention blocks and insertion positions in the ResNet-50 (see \autoref{fig:resnet}).
    You only need to add a single block of the proposed C-NL attention to ResNet-50 to outperform ResNet-101, and this only slightly affects the inference speed.}
\end{figure}

In both applications, it is mandatory for the robot to permanently be available to the users and to closely follow or guide them when desired.
To achieve this, the mobile robot has to track its users for a long time, even if they are occasionally out of sight for some time.
Therefore, person re-identification (re-id) plays a key role in such scenarios \cite{bellotto2007multisensor,alvarez2012feature,satake2013visual,eisenbach2015user,koide2016identification,an2017online,cosar2017volume,islam2019person,koide2019convolutional,algabri2020deep,algabri2022online,eirale2022human} either to provide robust features that can be used internally in a tracker for track continuation in case of temporal total occlusions \cite{muller2020multi} or to continue a track by re-identifying a person after a track is lost in a separate module \cite{wengefeld2016may}.

In recent years, ConvNet-based features have proven to be very reliable for non-biometric re-identification after the user has been unobserved for some time.
Therefore, these features can be used as visual cues in a probabilistic multi-person tracking framework \cite{muller2020multi}.
A person detector extracts cropped RGB images for all persons in the observed scene.
Next, single-shot re-identification features are extracted for these individuals.
The tracker then handles the temporal integration and fusion with other tracking inputs.

Fortunately, real-time deep-learning-based feature extraction on a mobile robot is no longer a problem due to the availability of specialized GPU devices such as the NVIDIA Jetson series.
However, inference speed still needs to be considered as these devices are typically shared for several neural-network-based classification or regression tasks.
This is in contrast to the current trend in person re-identification, where performance is improved at all costs, neglecting the significant slowdown during inference.
An example of this is the heavy use of attention blocks, especially non-local attention, which significantly slows down inference speed, as we will show in our experiments (see \autoref{sec:inferenceSpeed}).
We will demonstrate that we can improve the re-id performance with attention blocks without any significant slowdown in inference speed, if we find the right kind of attention blocks and a minimum set of appropriate positions in a ResNet-50 architecture.
Therefore, our contributions are as follows:

\vspace*{3pt}%
\begin{enumerate}
    \itemsep3pt
    \item We investigate the tradeoff between inference speed and re-id performance for different attention blocks, which is neglected in the current state of the art.
    \item We propose the new attention block C-NL that is faster than current attention blocks and performs better in a regime with few attention blocks.
    \item We perform extensive neural architecture search (NAS) to derive a set of rules for where and what type of attention should be integrated into a ResNet-50 to significantly improve re-id performance while only slightly affecting inference speed.
    \item We confirm that by integrating C-NL attention blocks into the ResNet-50 as specified in the derived set of rules, the re-id performance is also improved on a robotic dataset.
\end{enumerate}

\section{Related Work}

Nowadays, attention is widely used in computer vision.
To improve ConvNet-based single-shot person re-identification (re-id), mainly, three types of attention blocks are available:
channel-wise attention~\cite{hu2018squeeze}, %
spatial attention~\cite{wang2017residual}, %
and non-local attention~\cite{wang2018non}, %
with self-attention of the transformer architecture~\cite{vaswani2017attention} %
being a special form of the latter.

\begin{figure*}
    \centering
    ~~\includegraphics[height=7.5cm]{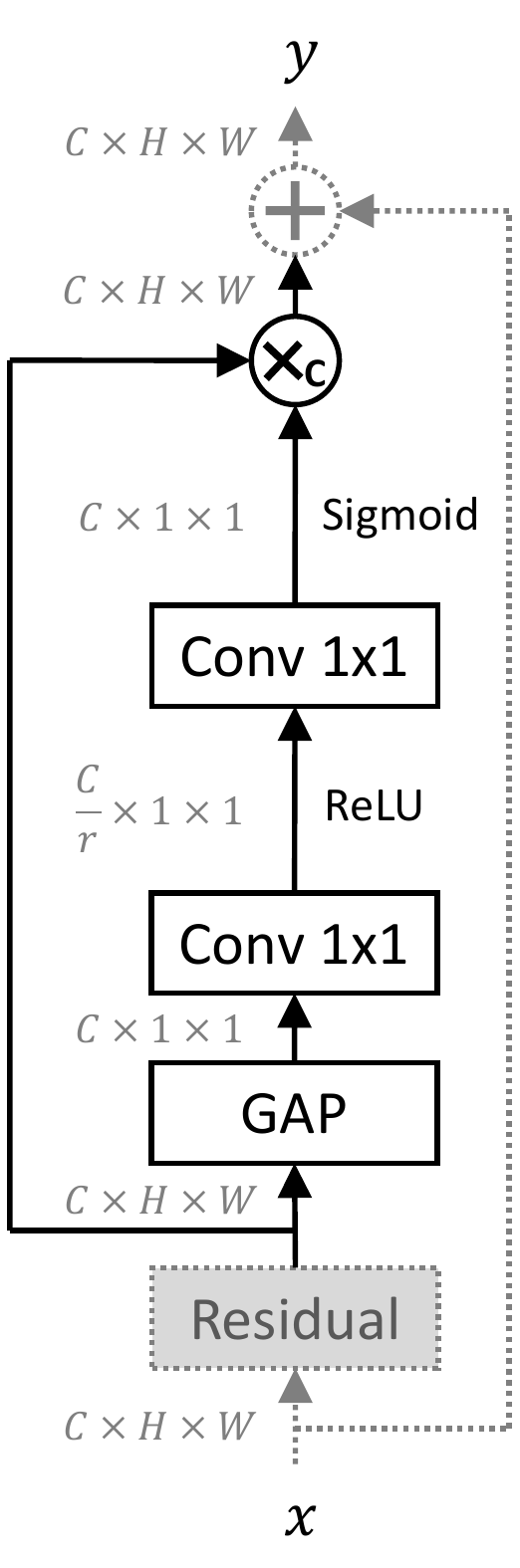}~~
    ~~\includegraphics[height=7.5cm]{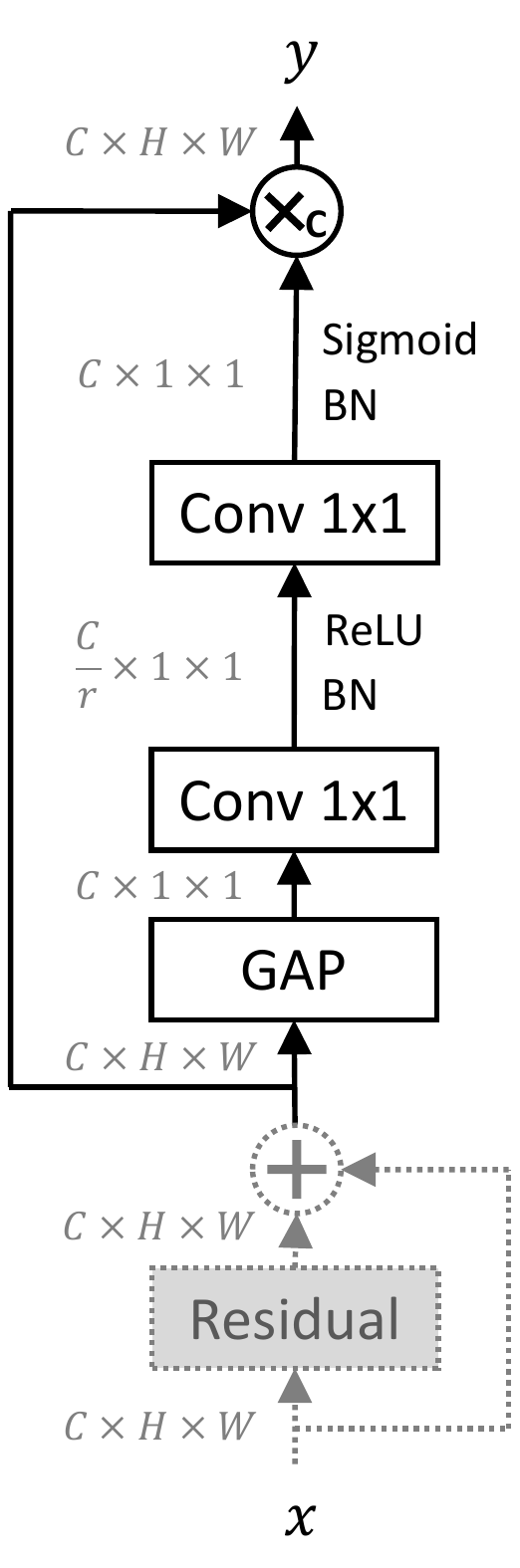}~~
    ~~\includegraphics[height=7.5cm]{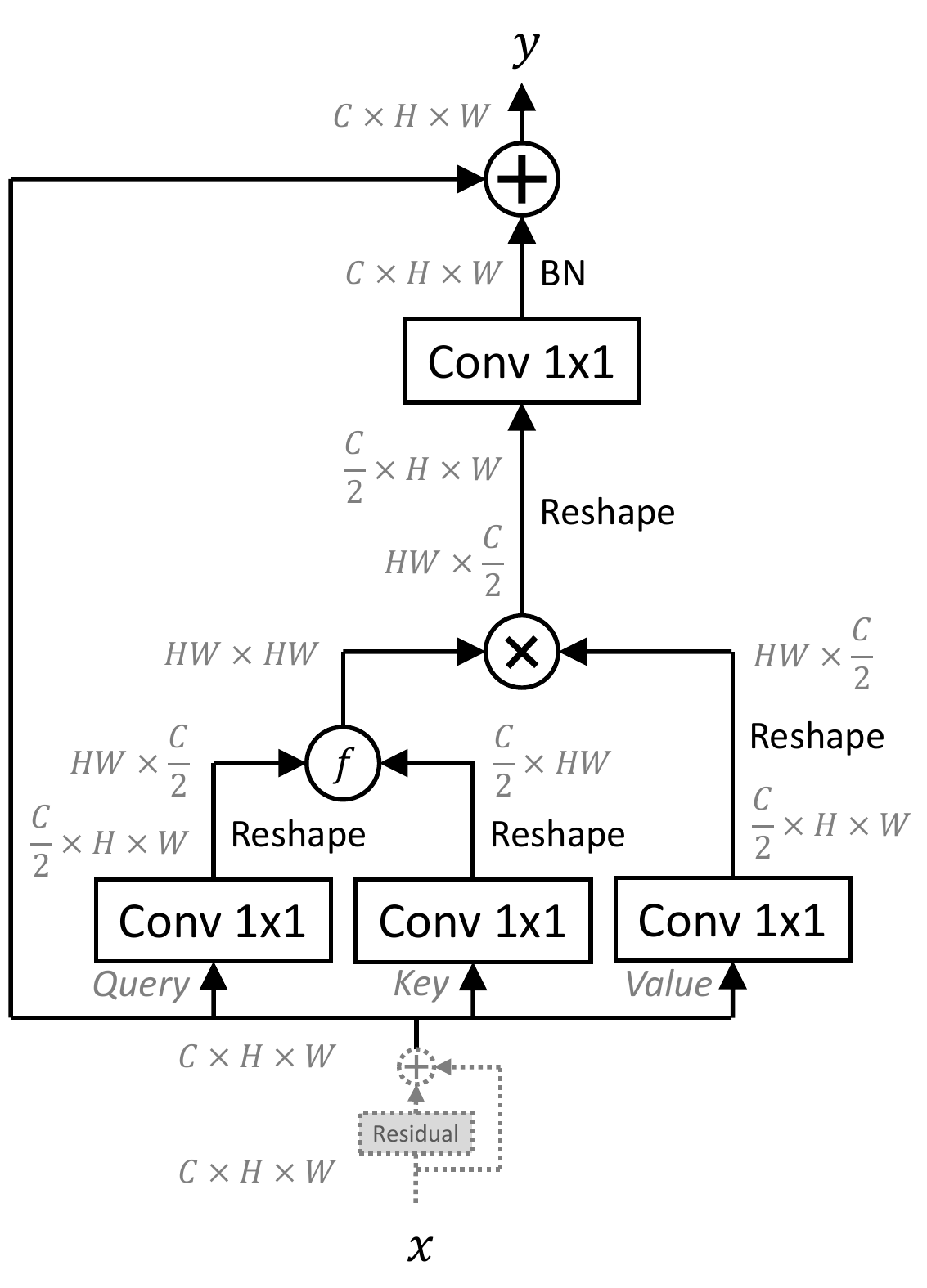}~~
    ~~\includegraphics[height=7.5cm]{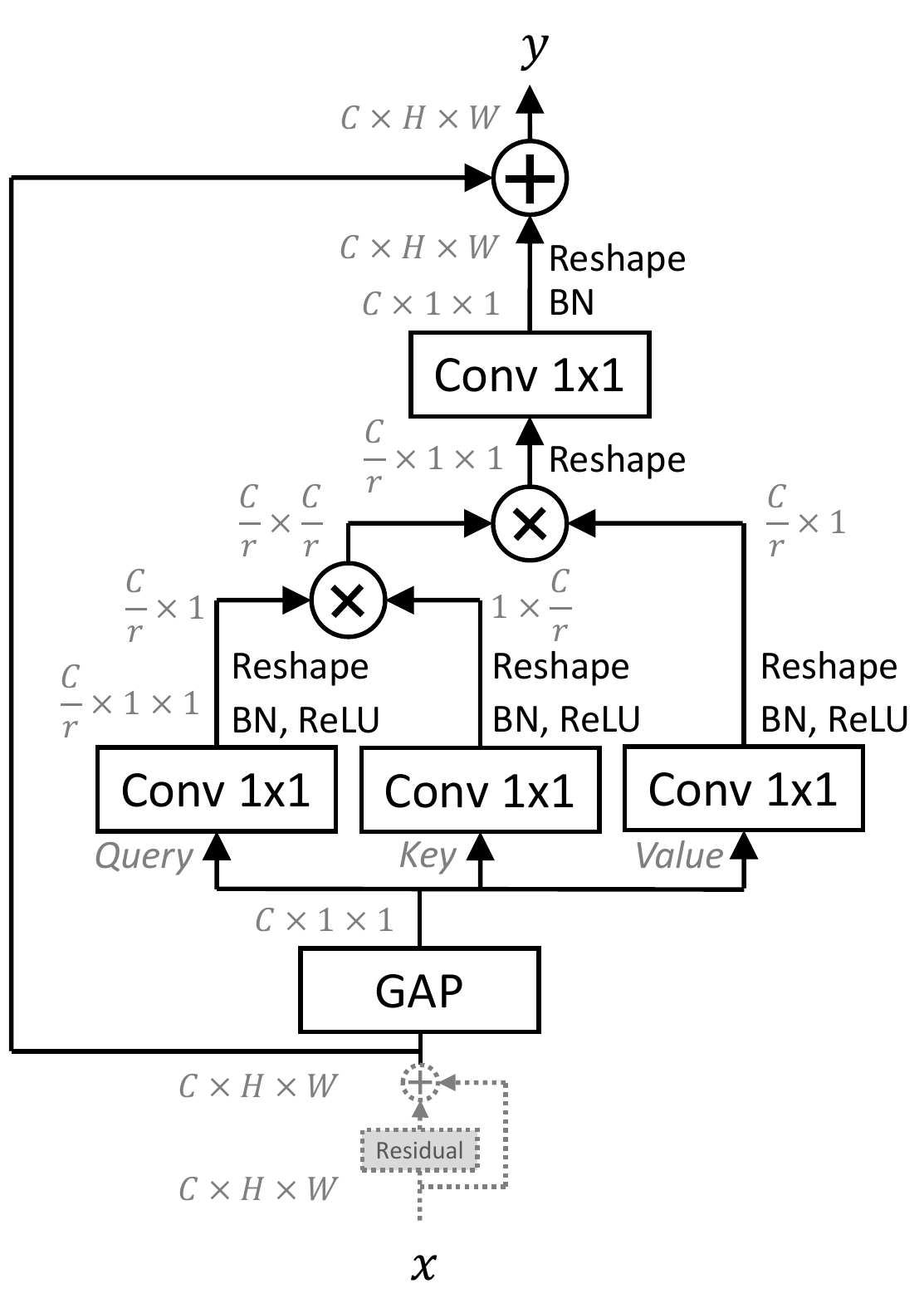}~~\\[-2mm]
    {\footnotesize(a) SE \cite{hu2018squeeze}\hspace{1.8cm}
    (b) HAC \cite{li2018harmonious}\hspace{2.5cm}
    (c) NL \cite{wang2018non}\hspace{4.1cm}
    (d) C-NL (proposed)\hspace{1cm}~}\\
    \caption{\label{fig:attentionBlocks}Layout of attention blocks: Channel-wise attention is realized by the (a) squeeze-and-excitation block \cite{hu2018squeeze} (SE) and the (b) harmonious attention channel-wise block \cite{li2018harmonious} (HAC). The prototype design of non-local attention \cite{wang2018non} (NL) is shown in (c). Our proposed block (C-NL) that is derived from these channel-wise and non-local attention blocks is shown in (d).}
\end{figure*}

\textit{Channel-wise attention}
is based on the principles of the squeeze-and-excitation block \cite{hu2018squeeze} (\autoref{fig:attentionBlocks}(a)).
First, the spatial resolution is reduced to $1{\times}1$ by global average pooling.
Then, the inter-channel correlation is modeled using a bottleneck in which the number of channels is reduced by $r$.
Finally, the output of the attention block, which is typically used as a per-channel weight, is normalized by a sigmoid activation.
Channel-wise attention based on these principles is widely applied for re-id \cite{li2018harmonious, yang2019attention, wu2021attention, chen2019self, ji2020attention, gong2021lag}.
Fully attention~\cite{wang2018mancs} is a variant of channel-wise attention for re-id that deliberately omits the global average pooling.
In \cite{chen2019mixed} fully attention was extended to higher-order statistics in the context of re-id.
We will show that channel-wise attention is cheap to compute, and should therefore be considered for a mobile robotic application where inference speed plays a key role.

\textit{Spatial attention}
is based on the network proposed in \cite{wang2017residual}.
First, the input volume is reduced to one channel while keeping the spatial resolution.
Then, the spatial resolution is reduced and subsequently restored to model the spatial correlation.
Finally, a sigmoid activation normalizes the output of the attention block, which is typically used as spatial weighting.
Spatial attention is also widely applied for re-id \cite{li2018harmonious, yang2019attention, wu2021attention, chen2019self, ji2020attention, zhou2019discriminative, li2020scanet}.
A variant that deviates from the design principles described above is the use of a foreground mask as spatial attention for re-id \cite{song2018mask, zhou2019foreground}.
This requires an additional network for extracting a foreground mask, which comes at an additional inference cost.
Spatial attention often does not perform that well on re-id benchmarks unless it is used in combination with other attention blocks.
Therefore, many re-id approaches combine spatial attention with channel-wise attention.
Channel-wise and spatial attention are computed in two branches that are either multiplied element-wise \cite{li2018harmonious, yang2019attention, wu2021attention} or concatenated at the channel dimension \cite{chen2019self, ji2020attention}.

\textit{Non-local attention}
aims at modelling global inter-pixel correlation.
Most often, it follows the design principles of the attention-module in the transformer architecture~\cite{vaswani2017attention} (\autoref{fig:attentionBlocks}(c)).
In a query and a key branch similarities of features are computed at different spatial positions which are then used to weight the features calculated in the value branch.
For re-id, non-local attention has been applied in this pure form \cite{ye2021deep, zhang2020relation}, with second-order statistics \cite{xia2019second}, and as self-attention \cite{chen2019abd}.
Non-local co-attention is also used to match person images of the probe and gallery \cite{zhang2019learning}.
The latter approach is therefore not relevant for our application of re-id features as tracking input, since we aim for extracting features for single person images.
Non-local attention achieves the best results in re-id benchmarks, but comes with high inference costs, as we will \mbox{show in our experiments.}

Besides these three main forms of attention blocks that are applicable for single-shot re-id on RGB images, there are also \textit{other forms of attention} applied for re-id.
In \cite{lan2017deep, chen2019self, zhang2020person} reinforcement learning is used to adapt attention weights.
In contrast, our focus is on learning the attention weights by error backpropagation.
Selecting proper semantic and soft-biometric attributes for re-id by attention has been addressed in 
\cite{zhang2020person, xu2018attention, tay2019aanet}.
To be able to utilize attributes would require an additional network for attribute extraction.
This is in contrast to our goal of affecting the inference speed only little.
Qian et al.~\cite{qian2019leader} use a multi-branch multi-scale architecture for person re-identification and train an attention block to select the best scale for an input.
Multi-scale approaches require significantly more operations, and therefore slow down inference, which is contrary to our goal.
Attention has also been applied to video-based re-id on streams of images~\cite{li2018diversity,si2018dual,karianakis2018reinforced,subramaniam2019co,fu2019sta,chen2019spatial,zhang2020multi,shim2020read,li2020relation,li2020temporal}.
These approaches primarily focus on temporal attention.
In addition, attention has been applied to re-id with other input modalities, such as infrared~\cite{ye2020dynamic} or depth images~\cite{haque2016recurrent}.
Both temporal attention and customized attention for other modalities are not the focus of the work presented in this paper.

\subsubsection*{Effect of integrating attention blocks on inference speed}
The drawback of adding attention blocks, namely slowing down inference, is completely neglected in the current state of the art.
However, fast inference is a prerequisite for real-time mobile robot applications.
Therefore, in this paper we address this issue.
We show that it is possible to integrate attention blocks at only few positions in a ResNet-50 to achieve similar performance to much deeper networks or computationally demanding attention blocks, while only marginally affecting the inference speed, if the correct positions can be identified.
Therefore, we perform extensive NAS to derive a set of rules to guide the integration.

\subsubsection*{Backbone}
We decided in favor of ResNet-50 as backbone for our experiments, since it is the de-facto standard for evaluating on person re-identification benchmarks.
Therefore, our results are easily comparable to the state of the art.
However, since we explore design spaces during NAS, the gained knowledge about integration of attention blocks is also transferable to other similar architectures \cite{radosavovic2020designing}.

\section{A Fast and Powerful Attention Block}
\label{sec:attentionBlocks}

After providing an overview of the state of the art in person re-identification (re-id) with architectures that incorporate attention blocks, we will now describe the attention blocks that we considered in our NAS.
Then, we derive a novel attention block that attempts to combine the advantages of these attention blocks in terms of inference speed and re-identification capabilities, as described in the following.

\subsection{Attention Blocks for Re-Identification}

\autoref{fig:attentionBlocks} shows the overall layout of channel-wise and non-local attention blocks that achieve the best re-id performance in benchmarks.
In \cite{li2018harmonious} (harmonious attention channel-wise, HAC), it has been shown that the performance (with channel-wise attention) improves most when attention blocks are inserted between residual blocks, as it has been shown first in \cite{wang2018non} for another type of attention applied to computer vision applications.
We visualize the modified position relative to the residuum in \autoref{fig:attentionBlocks}(b) (see \autoref{fig:attentionBlocks}(a) for the original position as proposed in the squeeze-and-excitation block \cite{hu2018squeeze}).
We also examined local attention multi-granularity network (LAG) \cite{gong2021lag}, but its attention block clearly underachieved in comparison to HAC.
Therefore, these results are not shown in the following experiments for reasons of clarity.

Also non-local attention blocks are placed between residual blocks (see \autoref{fig:attentionBlocks}(c)), which has been shown to be superior in \cite{wang2018non}.
We consider three non-local attention blocks, namely attention generalized mean pooling with weighted triplet loss (AGW) \cite{wang2018non} (applied for re-id in \cite{ye2021deep}), relation-aware global attention (RAGA) \cite{zhang2020relation}, and attentive but diverse (ABD) \cite{chen2019abd}.
All non-local attention blocks follow the same layout as shown in \autoref{fig:attentionBlocks}(c) with three branches, called query, key, and value in the transformer architecture \cite{vaswani2017attention}.
In all these three attention blocks, $f$ computes the matrix product of queries $q$ and keys $k$, with only the normalization of the result being different.
RAGA also comes with a further small difference, which translates into additional computational costs.

\begin{figure*}
    \begin{minipage}{0.81\linewidth}
        \includegraphics[width=\linewidth]{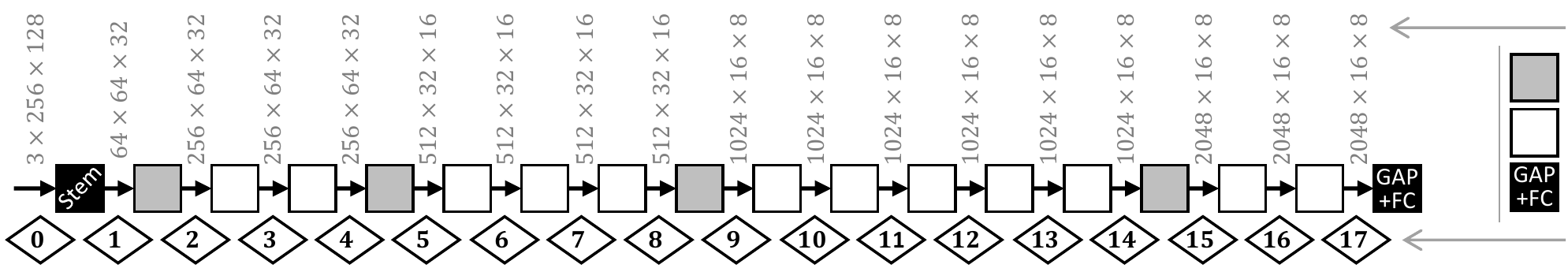}
    \end{minipage}
    \begin{minipage}{0.175\linewidth}
        \textit{\tiny Input tensor size for the attention}\\[-2.5mm]
        \textit{\tiny block at each possible position}\\[-1.4mm]
        \textit{\tiny Residual blocks with an $1{\times}1$}\\[-2.5mm]
        \textit{\tiny convolution in the skip connection}\\[-1.2mm]
        \textit{\tiny Normal bottleneck}\\[-2.5mm]
        \textit{\tiny residual blocks}\\[-1.0mm]
        \textit{\tiny Global average pooling (GAP) and fully}\\[-2.5mm]
        \textit{\tiny connected (FC) classification layer}\\[-1.0mm]
        \textit{\tiny Positions where attention blocks can}\\[-2.5mm]
        \textit{\tiny be inserted marked as diamonds}\\[-3.8mm]
    \end{minipage}
    \caption{\label{fig:resnet}ResNet-50 with positions where attention blocks can be inserted marked as diamonds. Applying attention blocks directly on the input is somewhat questionable and has shown to not improve the performance in our experiments. All other 17 positions between the initial stem and the final GAP+FC should be considered. Residual blocks with an $1{\times}1$ convolution in the skip connection (gray) mark the beginning of each of the four stages of the ResNet-50.}
\end{figure*}

\subsection{Deriving a Novel Attention Block}

Both channel-wise and non-local attention blocks provide advantages for re-identification, which we attempt to combine in our novel attention block as described below.
In our experiments, we show that non-local attention blocks slow down inference speed too much to be worth considering for a mobile robotic application.
Therefore, our novel block should be designed like the HAC attention block \cite{li2018harmonious}.
Its low inference costs are a result of the early reduction of the spatial resolution by global average pooling.
The benefit of non-local attention blocks are, however, the three-branch layout to appropriately weight the features in the value branch to estimate global correlations.
Therefore, our block uses an early global average pooling to reduce computational costs and then follows the three-branch design of the non-local attention blocks to model the inter-channel correlation in a non-local way.
\autoref{fig:attentionBlocks}(d) shows the derived design of our channel-wise non-local attention block (C-NL).
The three-branch layout comes with minor computational costs in this setting as a result of the matrices of reduced size that need to be multiplied (see matrix dimensions shown in gray in \autoref{fig:attentionBlocks}(c, d)).
Therefore, this design leads to low computational costs and achieves a high re-id performance, as we show in the following experiments.

\section{How Much Attention Do We Need for Person Re-Identification?}

In what follows, we will first address the knowledge gap regarding the extent to which attention blocks slow down inference speed.
Then, we will analyze how many attention blocks we really need to significantly improve person re-identification (re-id), by performing extensive NAS over the design space of attention blocks, positions of integration, and hyperparameters.
We take inspiration by the Designing Network Design Spaces NAS approach \cite{radosavovic2020designing} that showed how the design space can gradually be reduced and, as a consequence, a set of design rules can be derived.
We primarily reduce the design space regarding positions of integration to derive a set of rules of where attention blocks should be added to get a large improvement with low computational cost.

\subsection{Experimental Setup}

As basis for our experiments, we use the strong baseline and training setup of \cite{luo2019bag} consisting of a modified ResNet-50 \cite{he2016deep}, which we also deploy in real-time on our mobile robots, in combination with common augmentation and training strategies.
For benchmarking, we decided in favor of one of the most popular datasets, namely Market-1501 \cite{zheng2015scalable}, and used the evaluation protocol of \cite{ye2021deep}.
We could also benchmark on DukeMTMC-reID \cite{ristani2016performance} or CUHK03-NP \cite{li2014deepreid}, but instead we decided in favor of a robotic dataset, for which we report results in Sec.~\ref{sec:robotic_dataset}, to show generalization ability.

Note, that better results can be achieved on Market-1501 with more elaborate baselines.
But these baselines come with architectural modifications, which result in higher computational costs.
This is contrary to our goal of fast inference on a robot.
Also by integrating several time frames, like in \cite{wang2019spatial}, or other costly techniques, better benchmark results could be achieved, but this is again contrary to our goal of deploying these networks in real-time on a mobile robot.

During NAS, all hyper-parameters are tuned, such that the reported results are trustworthy and not a fragment of the choice of hyper-parameters.

\subsection{Analysis of Inference Speed of Attention Blocks for Person Re-Identification}
\label{sec:inferenceSpeed}

First, we measure the inference speed of attention blocks that have been proposed or applied for re-id as described in \autoref{sec:attentionBlocks}\footnote{For implementation of the attention blocks, where available, we used publicly available code provided by the authors: AGW, evaluation code -- {\tiny\url{https://github.com/mangye16/ReID-Survey}}, RAGA -- {\tiny\url{https://github.com/microsoft/Relation-Aware-Global-Attention-Networks}}, ABD -- {\tiny\url{https://github.com/VITA-Group/ABD-Net}}, LAG -- {\tiny\url{https://github.com/SWJTU-3DVision/LAG-Net}}}.
We add the different attention blocks at single positions in the ResNet-50.
The diamonds in \autoref{fig:resnet} show the locations where attention blocks can be integrated.
We measure the inference speed on a Jetson AGX Xavier (Jetpack 4.6, TensorRT 8.0.1.6, Cuda 10.2, 16 bit floating point precision), which is a special device typically used for deep-learning computations on a mobile robot.
Since for each image we need to extract features for all persons in the camera, we chose a batch size of 16 for the measurement of inference speed.
With this batch size, we would be able to extract features for 16 persons in the field of view simultaneously, which we consider as a good worst-case estimate for crowded scenes, like in hospital hallways.
We report the number of batches that can be processed per second during inference.
The reported results are the average over 10000 processed batches of the Market-1501 benchmark dataset \cite{zheng2015scalable} with cropped and scaled RGB person images of size $256{\times}128$.
However, the inference speed does not depend on the dataset, but only on the input image size, which is identical in all our experiments in this paper.

\begin{figure}[t!]
    \centering
    \begin{minipage}{0.83\linewidth}
        \includegraphics[width=\linewidth]{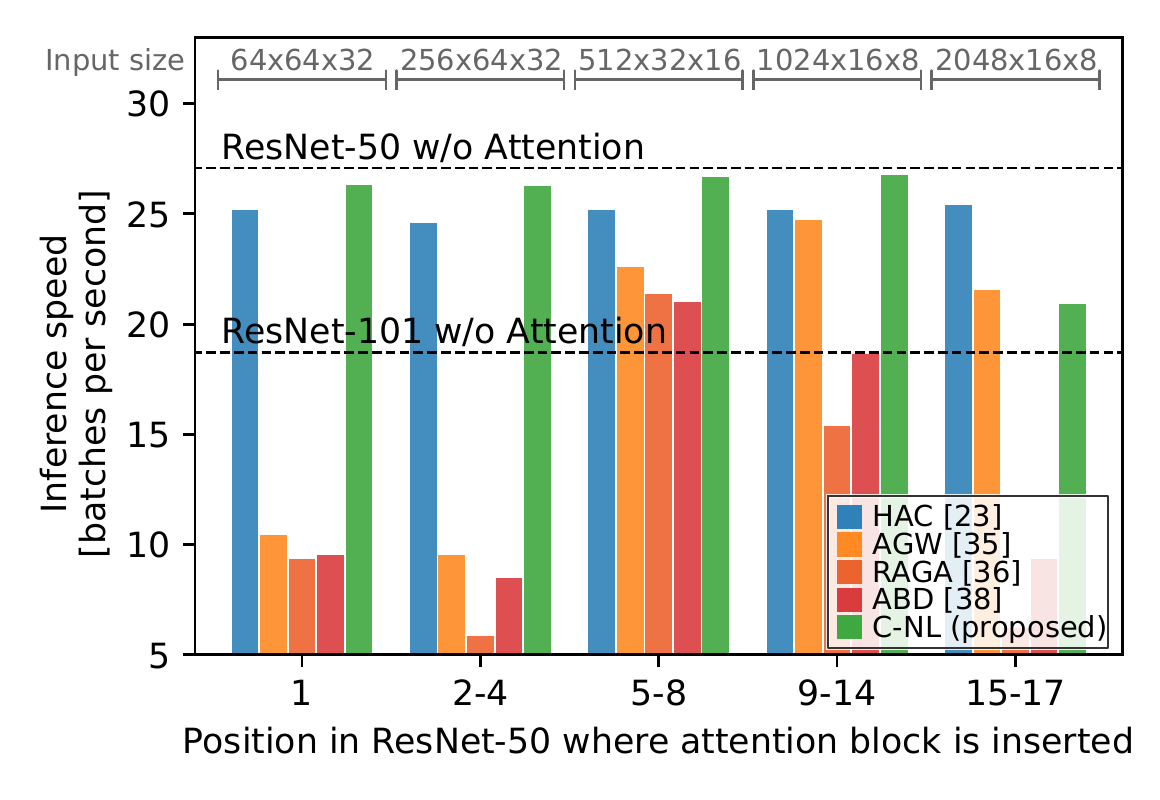}
    \end{minipage}
    \begin{minipage}{0.155\linewidth}
        \includegraphics[width=\linewidth]{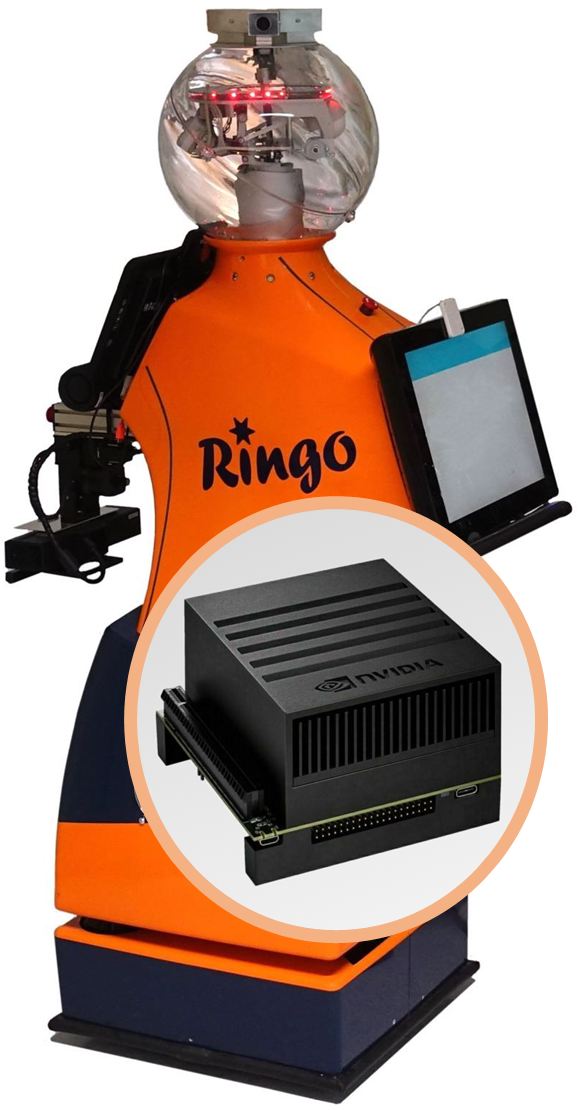}\\[-1mm]
        \textit{\tiny Jetson AGX}\\[-2.5mm]\textit{\tiny Xavier for deep}\\[-2.5mm]\textit{\tiny  learning com-}\\[-2.5mm]\textit{\tiny putations on}\\[-2.5mm]\textit{\tiny our rehab com-}\\[-2.5mm]\textit{\tiny panion robot}\\[3mm]
    \end{minipage}
    \caption{\label{fig:speedPosition}Inference speed on a Jetson AGX Xavier for integrating single attention blocks at different positions measured by number of batches processed per second. The input tensor size for each attention block at the respective positions is listed at the top of the figure.}
\end{figure}

\begin{figure*}
    \centering
    \includegraphics[width=\linewidth]{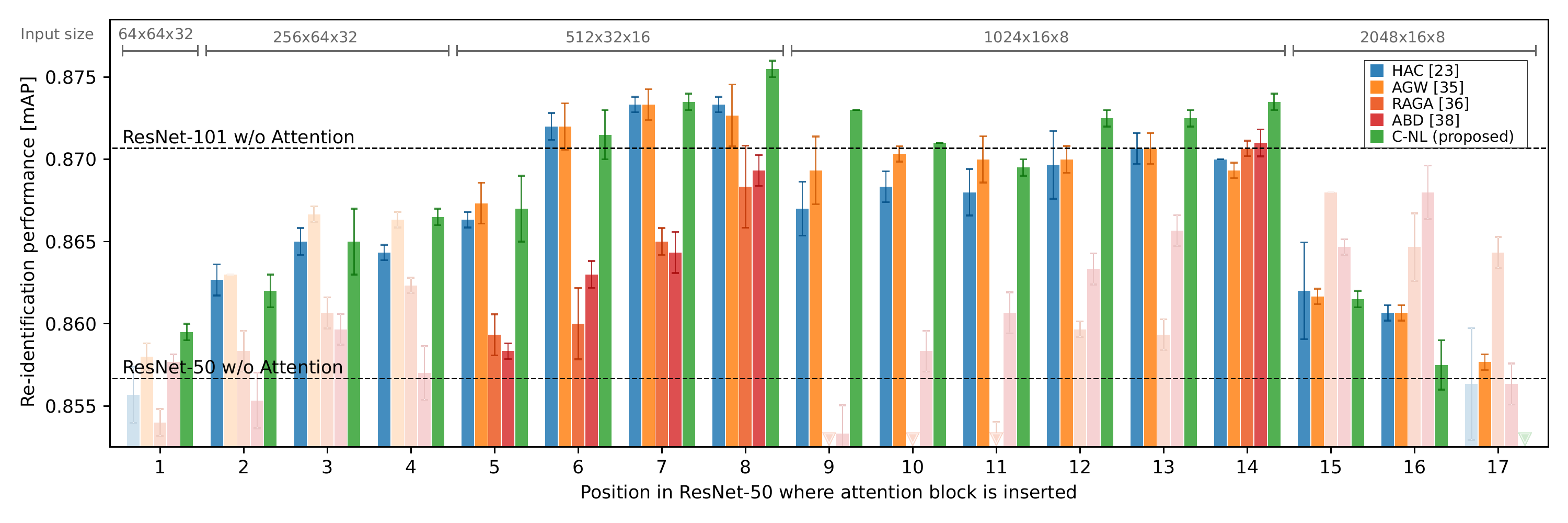}
    \caption{\label{fig:mapPosition}Re-id performance measured by the mean average precision (mAP) on the Market-1501 benchmark dataset \cite{zheng2015scalable} for different attention blocks integrated at a single position in the ResNet-50 (as displayed in \autoref{fig:resnet}).
    Downwards pointing triangles symbolize that the bar is below the displayed range of values.
    Results for positions where inference with attention blocks is slower than ResNet-101 but does not match its performance, or where its performance is even worse than ResNet-50, are shown in light colors to symbolize that these configurations should not be considered.}
\end{figure*}

\autoref{fig:speedPosition} shows the inference speed for the attention blocks as a function of the single position where the individual attention block is integrated.
Note that the inference speed depends only on the size of the input tensor for the attention block (listed at the top of the figure) and is thus identical for several positions.
As reference, we show the inference speed of the ResNet-50 without any attention block and of the ResNet-101 (without attention) that is twice as deep.
We can see that channel-wise attention blocks shown in blue and green (HAC \cite{li2018harmonious}, C-NL) are faster to compute than non-local attention blocks shown in orange to red (AGW \cite{ye2021deep}, RAGA \cite{zhang2020relation}, ABD \cite{chen2019abd}).
A ResNet-50 with a single non-local attention block at an early position (1--4) is significantly slower than a ResNet-101.
Thus, these configurations should not be considered for robotics applications.
Furthermore, at late positions (9--17) only AGW \cite{ye2021deep} can cope with the inference speed of channel-wise attention blocks to some degree.
Also at middle positions (5--8) channel-wise attention is faster to compute than non-local attention.
Comparing the channel-wise attention blocks, for most positions (1--14) the proposed C-NL block achieves a higher inference speed than HAC \cite{li2018harmonious}.
To conclude, considering only inference speed so far, channel-wise attention is preferable over non-local attention in robotic applications.

\subsection{How Many Attention Blocks Do We Really Need?}

Next, we analyze, how many attention blocks we need to significantly improve the re-id performance.
Therefore, inspired by \cite{radosavovic2020designing}, we performed NAS by progressively reducing the design space over attention blocks, positions of integration, and hyperparameters regarding positions of integration while considering computational costs as measured above.
\autoref{fig:mapPosition} shows the re-id performance when attention blocks are inserted at a single position measured by the mean average precision (mAP) on Market-1501 \cite{zheng2015scalable}.
Furthermore, the plot shows mean and standard deviation over three training runs.

As we can see, even a single attention block integrated at a suitable position can top the performance of the ResNet-101 that is twice as deep as the baseline.
Furthermore, we can see that channel-wise and non-local attention perform very similar in this regime of few attention blocks.
By considering the lower inference costs, we clearly should prefer channel-wise attention over non-local attention.

\subsection{Where Should We Add Attention Blocks?}

Now that we have seen that adding a single attention block is a powerful approach, we like to investigate whether combinations of computationally effective channel-wise attention blocks at different positions can further improve the performance.
From the extensive NAS with plenty of combinations of positions and types of attention, as part of the progressive reduction of the design space, we were able to derive the following set of rules to improve the re-identification performance most by using as few as possible attention blocks:

\vspace*{3pt}%
\begin{enumerate}
    \itemsep3pt
    \item Select positions where the single-position performance is good (see \autoref{fig:mapPosition}). Most often, positions at the end of each stage seem to perform slightly better than earlier positions in each stage.
    \item The distance between selected positions should be large. Positions in different ResNet stages should be preferred.
    \item A combination of different types of attention (channel-wise, non-local) performs slightly better than single-type combinations (but is not worth it for robotic applications due to the computational cost of non-local attention blocks).
\end{enumerate}

\begin{table}[t!]
    \caption{Re-id performance on the Market-1501 benchmark dataset~\cite{zheng2015scalable} and inference speed for attention blocks at single and multiple positions using cross entropy and circle loss~\cite{sun2020circle} respectively.}
		\vspace*{-2mm}%
    \centering
    \setlength{\tabcolsep}{1.0pt}
    \begin{tabular}{cccc}
        \hline
        & & Re-id performance & Inference speed \\
        Attention & Loss function & [mAP] & [batches/second] \\
        \hline
        HAC \cite{li2018harmonious} @ 8 & cross entropy & 0.8733 $\pm$ 0.0006 & 25.17 \\
        \rowcolor{lightgray!50!white!80} AGW \cite{ye2021deep} @ 7 & cross entropy & 0.8733 $\pm$ 0.0012 & 22.59 \\
        C-NL @ 8 & cross entropy & 0.8755 $\pm$ 0.0007 & 26.70 \\
        \hline
        \rowcolor{lightgray!50!white!80} HAC \cite{li2018harmonious} @ 8, 14 & cross entropy & 0.8773 $\pm$ 0.0006 & 25.06 \\
        AGW \cite{ye2021deep} @ 8, 14 & cross entropy & 0.8805 $\pm$ 0.0014 & 21.09 \\
        \rowcolor{lightgray!50!white!80} C-NL @ 8, 14 & cross entropy & 0.8788 $\pm$ 0.0007 & 25.91 \\
        \hline
        HAC \cite{li2018harmonious} @ 6, 8, 14 & cross entropy & 0.8780 $\pm$ 0.0010 & 24.93 \\
        \rowcolor{lightgray!50!white!80} AGW \cite{ye2021deep} @ 6, 8, 14 & cross entropy & 0.8815 $\pm$ 0.0008 & 19.88 \\
        C-NL @ 6, 8, 14 & cross entropy & 0.8806 $\pm$ 0.0007 & 25.89 \\
        \hline
        \rowcolor{lightgray!50!white!80} HAC \cite{li2018harmonious} @ 6, 8, 14 & circle loss \cite{sun2020circle} & 0.8897 $\pm$ 0.0012 & 24.93 \\
        AGW \cite{ye2021deep} @ 6, 8, 14 & circle loss \cite{sun2020circle} & 0.8916 $\pm$ 0.0006 & 19.88 \\
        \rowcolor{lightgray!50!white!80} C-NL @ 6, 8, 14 & circle loss \cite{sun2020circle} & \textbf{0.8916} $\pm$ 0.0005 & \textbf{25.89} \\
        \hline
    \end{tabular}
    \label{tab:attentionCombinations}
		\vspace*{-4mm}%
\end{table}

\vspace*{3pt}%
The best results of the NAS for each type of attention block was achieved for a pair of attention blocks at positions 8 and 14, and for the triplet of attention blocks at positions 6, 8, and 14.
Using the same type of attention block at more than three positions did not improve the results further.
\autoref{tab:attentionCombinations} lists the best results for attention blocks at multiple positions.
Furthermore, \autoref{fig:mapVsSpeed} shows the re-id performance versus inference speed in a graph.
We can see that for the channel-wise attention blocks HAC and C-NL, the inference speed is affected only slightly.
Inference with the proposed C-NL at positions 6, 8, and 14 is only 1.706$\,$ms slower (ca. 5\%) than the ResNet-50 baseline without any attention blocks (38.623$\,$ms vs. 36.917$\,$ms).
Therefore, while just 1.196 fewer batches can be processed per second, the re-identification performance far exceeds that of ResNet-101 (mAP 0.8806 $\pm$ 0.0007 vs. 0.8707 $\pm$ 0.0006), which is twice as deep and takes 53.454$\,$ms to process a batch.

\subsection{Can the Performance Be Improved Further?}
\label{sec:lossFunction}

In \cite{aganian2021revisiting}, an alternative way of improving the re-identification performance was examined.
By using a large-margin loss function, the recognition rate is increased without affecting the inference speed.
We also apply the best performing loss function, namely circle loss \cite{sun2020circle}.
In \cite{aganian2021revisiting}, circle loss achieved a mAP of 0.882 on the Market-1501 dataset using the strong baseline \cite{luo2019bag} setup as in our experiments.
We observed, that the combination of a large-margin loss function and few attention blocks at suitable positions complement each other very well and boosts the performance further (see \autoref{tab:attentionCombinations}).
The proposed C-NL attention block matches the performance of the much more costly AGW non-local attention block, while being faster during inference (highlighted in bold).

\section{Experiments on a Robotic Dataset}
\label{sec:robotic_dataset}

To examine, whether our improvements by using attention blocks at few positions can also be achieved on robotic data, we conducted an experiment on an extended version of the ROREAS dataset, which was recorded with a mobile robot following a user in a clinical environment \cite{eisenbach2015user}.
We will refer to this extended version as ROREAS+ in the following.

The ROREAS+ dataset differs significantly from typical re-id benchmark datasets, like Market-1501 \cite{zheng2015scalable}, DukeMT\-MC-reID \cite{ristani2016performance}, or CUHK03-NP \cite{li2014deepreid}.
These re-id benchmark datasets have been obtained from footage taken from outdoor surveillance cameras mounted at several meters height.
In contrast, our target robotic platforms, as in \cite{eisenbach2015user}, operates inside buildings and the camera is mounted at eye level.
This results in completely different images, which differ in perspective, occlusions, lighting conditions, and contrast from the typical benchmark datasets.
Through our evaluation on the ROREAS+ robotics dataset, which follows the protocol of \cite{ye2021deep} as in all experiments in this paper, we should get a clear picture regarding how much attention helps to improve person re-identification in a real-world robotics application scenario.
Due to data protection laws, we are not allowed to show pictures, but only report benchmarking results.

\begin{table}[b!]
		\vspace*{-4mm}%
    \caption{Re-id performance on the robotic dataset RO\-RE\-AS+ using a ResNet-50 with and without attention.}
		\vspace*{-2mm}%
    \centering
    \setlength{\tabcolsep}{2.6pt}
    \begin{tabular}{ccccc}
        \hline
        & Dataset pre-taining & Dataset fine-tuning & Attention & mAP \\
        \hline
        (1) & Market-1501 \cite{zheng2015scalable} & --- &  --- & 0.6630 \\
        \rowcolor{lightgray!50!white!80} (2) & Market-1501 \cite{zheng2015scalable} & --- & C-NL & \textbf{0.6684} \\
        \hline
        (3) & Market-1501 \cite{zheng2015scalable} & ROREAS+ & --- & 0.7551 \\
        \rowcolor{lightgray!50!white!80} (4) & Market-1501 \cite{zheng2015scalable} & ROREAS+ & C-NL & \textbf{0.7658} \\
        \hline
    \end{tabular}
    \label{tab:roboticResults}
\end{table}

The ROREAS+ dataset contains 421 different persons.
In order to get a realistic setup, out of 421 recorded persons, we have chosen only 50 persons with the largest amount of recorded images as training data, the rest is taken as test data to get a representative test set with many different (but potentially similarly looking) persons included, where attention can make the difference.
Recording many views for 50 people with a mobile robot can already be a challenging and potentially time-consuming task in some scenarios, but we found this number to be a good trade-off between the amount of work needed to record the data and the amount of data available for a deep-learning approach.
Therefore, 34012 images of 50 identities are available for training, while the test set is split into 945 query images and 2520 gallery images of 371 identities.

Due to the limited amount of training data, we must use transfer learning to profit from pre-training on larger re-id benchmark datasets.
Especially the small amount of different identities can be an issue, otherwise.
For transfer learning, we first pre-train on Market-1501 \cite{zheng2015scalable} and then use a two-step fine-tuning approach \cite{chen2018deep}, where first the randomly initialized classification layer is fine-tuned on our target dataset ROREAS+, followed by the fine-tuning of all network weights.
For training, we follow the strong baseline setup \cite{luo2019bag}.
As loss function, we employ circle loss, as it showed to be beneficial in our previous experiment.
Following our best results, we added the best performing attention block C-NL at positions 6, 8, and 14.

In \autoref{tab:roboticResults}, we report the best results over three runs, since this would be the configuration that we would select for later application.
First, we could use a ResNet-50 for feature extraction that was trained on re-id benchmark data only, as in our previous experiments, without any fine-tuning on ROREAS+.
We can see, that using a ResNet-50 for feature extraction that was never trained on robotic data (rows 1, 2) performs worse, both with and without attention, compared to the corresponding results with fine-tuning on ROREAS+ (rows 3, 4).

In both cases -- with or without finetuning -- C-NL noticeably improves the ability to re-identify persons. The standard deviation over three runs is $\sigma = 0.006$. Therefore the results with attention are ca. $1.5 \sigma$ better than without attention.

\section{Conclusion}
In the state of the art on person re-identification, inference cost is widely neglected, and adding attention blocks at only few positions was not considered so far.
We have shown that adding a channel-wise attention block at a single position in a ResNet-50 is sufficient to perform better in person re-identification than a ResNet-101.
Therefore, the common practice of adding attention to each residual block should be questioned.

In addition, due to the progressive reduction of the design space during neural architecture search, we were able to derive a set of rules for where attention blocks should be integrated in a ResNet architecture.
By integrating the proposed attention block C-NL at three positions in a ResNet-50 derived by the NAS, we improved the mAP for the Market-1501 dataset, surpassing the performance of the ResNet-101 by a large margin, while slowing down the inference by only 5\%, which is marginal compared to the 45\% slowdown caused by doubling the depth.
Combining this architectural modification with a large-margin loss function further improved the re-identification ability.
Compared to HAC block \cite{li2018harmonious}, which slows down inference by 9\% (nearly twice as much as C-NL) and performs $1.6\sigma$ worse in this setup, the modifications of the C-NL attention block building on the channel-wise attention block setup of HAC proved effective.

Finally, we confirmed the performance gain for robotic data when transfer learning with pre-training on a re-id benchmark dataset is applied.
The improved mAP shows that few attention blocks in the architecture enable the extracted re-id features to better discriminate individuals in difficult cases where individuals look very similar.
This will help the robot to better keep track of its users and will consequently improve clinical applications where a mobile robot needs to follow or guide its user.

In future work it would be interesting to apply the proposed approach to neural architectures designed for mobile devices that have shown fast inference speed on Jetson hardware, such as ShuffleNet v2 \cite{ma2018shufflenet} or MobileNet v3 \cite{howard2019searching}.
We are curious if the same design rules apply and if a single channel-wise attention block is sufficient for these architectures as well.

\bibliographystyle{IEEEtran}
\bibliography{sec/references} 

\end{document}